# Tracing Cultural Diachronic Semantic Shifts in Russian Using Word Embeddings: Test Sets and Baselines

**Fomin V.** (wadimiusz@gmail.com),
**Bakshandaeva D.** (dbakshandaeva@gmail.com),
**Rodina Ju.** (julia.rodina97@gmail.com)

National Research University Higher School of Economics, Moscow, Russia

**Kutuzov A.** (andreku@ifi.uio.no)

University of Oslo, Oslo, Norway

The paper introduces manually annotated test sets for the task of tracing diachronic (temporal) semantic shifts in Russian. The two test sets are complementary in that the first one covers comparatively strong semantic changes occurring to nouns and adjectives from pre-Soviet to Soviet times, while the second one covers comparatively subtle socially and culturally determined shifts occurring in years from 2000 to 2014. Additionally, the second test set offers more granular classification of shifts degree, but is limited to only adjectives.

The introduction of the test sets allowed us to evaluate several well-established algorithms of semantic shifts detection (posing this as a classification problem), most of which have never been tested on Russian material. All of these algorithms use distributional word embedding models trained on the corresponding in-domain corpora. The resulting scores provide solid comparison baselines for future studies tackling similar tasks. We publish the datasets, code and the trained models in order to facilitate further research in automatically detecting temporal semantic shifts for Russian words, with time periods of different granularities.

# КУЛЬТУРНО ОБУСЛОВЛЕННЫЕ ДИАХРОНИЧЕСКИЕ СДВИГИ ЗНАЧЕНИЙ СЛОВ В РУССКОМ ЯЗЫКЕ: ТЕСТОВЫЕ СЕТЫ И БАЗОВЫЕ ДИСТРИБУТИВНЫЕ АЛГОРИТМЫ

**Фомин В.** (wadimiusz@gmail.com),
**Бакшандаева Д.** (dbakshandaeva@gmail.com),
**Родина Ю.** (julia.rodina97@gmail.com)

НИУ Высшая школа экономики, Москва, Россия

**Кутузов А.** (andreku@ifi.uio.no)

Университет Осло, Осло, Норвегия

В статье представлены размеченные вручную датасеты для тестирования систем автоматического отслеживания изменений значений слов в русском языке. Датасеты взаимодополняемы: первый из них посвящён семантическим изменениям прилагательных и существительных в русском языке советского периода в сравнении с периодом до 1918 года, тогда как второй связан с тонкими семантическими изменениями, происходившими между годами с 2000 по 2014. Кроме того, второй датасет предлагает гранулярную классификацию степени изменения, однако содержит только прилагательные.

Создание этих датасетов позволило оценить качество работы нескольких распространённых алгоритмов определения семантических сдвигов, многие из которых ранее не применялись к русскому языковому материалу. Все использованные алгоритмы основаны на дистрибутивных моделях, обученных на соответствующих корпусах, саму задачу оценки алгоритмов можно отнести к задачам классификации. Итоговые оценки могут считаться базовым уровнем (baseline) для сравнения с будущими подходами. Датасеты, программные реализации использованных алгоритмов и обученные дистрибутивные модели выложены в открытом доступе. Мы надеемся, что это поможет будущим исследованиям в области автоматического отслеживания семантических сдвигов в русском языке с различной временной гранулярностью.

**Ключевые слова:** дистрибутивная семантика, диахронические семантические сдвиги, изменения языка во времени, датасеты

## 1. Introduction

In any language, words undergo semantic shifts over time, that is, they acquire new meanings and lose the old ones. The sources and nature of these shifts can vary. For example, words can completely change what they mean (cf. the Russian adjective 'красный' shifting from '*beautiful*' to '*red*'[1]). Sometimes, these shifts are linked to global processes of language development, and the granularity of the involved time spans is often very large-scale: we are talking about centuries or more.

Another type of diachronic semantic shifts is related to changes which occur on comparatively small time spans (decades or even years). The traditional classification in [25] assigns them to the category of *substitutions*: changes that have non-linguistic causes, for example that of technological progress. Such shifts often do not radically change the core meaning of a word, but instead significantly restructure the sets of typical associations which the word triggers in the speakers' minds. Cf., for example, the metonymical changes in the semantic structure of the Russian word 'Болотная [площадь]' '*bolotnaya [square]*', when after the 2012 mass protests, its meaning has widened to include not only the toponym in Moscow, but also the protest movement itself. These 'cultural shifts' are sometimes defined as socially and culturally determined changes in the people's associations for a given word [8].

This definition plays well with the so called distributional hypothesis [6]. In the present paper, we treat word meaning (or lexical semantics) as a **function of words' contexts in natural texts**. It is important that there exist many other linguistic theories and definitions of what meaning is: for example, one can postulate that the word meaning equals to the definition of this word in a well-established dictionary. We do not claim superiority of any of these theories; however, in our work, we stick to the distributional one. Thus, for us semantic shifts are primarily changes in the word's typical contexts and associations. The task of automatic discovery of diachronic semantic shifts using data-driven methods (especially distributional semantic models) is becoming increasingly popular in contemporary computational linguistics: see [18] and [26] for only a few of the recent surveys. However, to the best of our knowledge, there is only one publication applying these methods to Russian material [16]; it is behind the paywall, and thus not publicly accessible.

The presented paper aims to establish at least some foundation for further studies in automatically tracing temporal semantic shifts (of different granularities) for Russian. Correspondingly, our main **contributions** are as follows:

1. we release a manually annotated dataset of socially and culturally determined semantic shifts in Russian adjectives, based on news texts published from 2000 to 2014, covering many important social and political events;
2. we re-package in a more machine-readable and coherent form the dataset of Russian nouns and adjectives which shifted their meaning in the Soviet period [16];
3. several well-established algorithms of tracing diachronic semantic shifts using word embeddings are tested on the aforementioned datasets; this provides publicly available baseline scores for this task, when applied to Russian.

---

[1] Vasmer's Etymological Dictionary of the Russian Language, 1986

The rest of the paper is structured as follows. In **Section 2** we put our work in the context of the related research, both globally and in Russia. **Section 3** describes the process of annotating the presented gold standard datasets. **Section 4** introduces the corpora we employed to test our hypothesis and to train distributional semantic models, in its turn described in **Section 5**. We evaluate the algorithms of semantic shifts detection in **Section 6**, and in **Section 7** we discuss the results and conclude.

## 2. Previous work

Linguistics has a long story of studying diachronic semantic shifts. Just like words change, methods and sources used to trace changes in word's meaning and goals set in such research also evolve. The hand-picking approach [27] was historically the first, but then, with the expansion of corpus and computational linguistics, distributional semantics in particular, the data-driven approach was brought to the foreground. Time spans used to compare word's usage were initially as large as centuries [24], but then have begun to shrink to decades [7] and even years [14, 19]. Word embedding models, which appeared to be the most efficient tool in the field, also vary. For example, there are models that are trained separately on the corresponding time spans and aligned using specific methods, such as 'local anchors' [30] or orthogonal Procrustes transformation [9]. In other cases, word embeddings are trained jointly across all time periods, like in dynamic models of [1] and [28]. In yet other approaches, the models are trained incrementally, so that the model for every next time period is initialised with the embeddings from the previous period [14]. As for the data, in the beginning extensive and inclusive corpora were used, which primarily contain book texts [7, 12], but later researchers have taken interest in more specific corpora, for instance, of newspapers [3, 10, 28].

Unfortunately, in most cases only British and American English material is analysed, while there are much less studies dedicated to other languages [26]. As postulated in [18], the lack of studies in which existing methods are applied to other languages apart from English is among top priority open challenges in the field of detecting diachronic semantic shifts.

For temporal semantic shifts studies in Russian, [4] provides an inspiring linguistic foundation. Examining the corpus[2] (exploring the contexts in which the particular word appears and counting the number of word's uses for each period) gave insights about transformations of meaning which words undergo. Unfortunately, the words were once again hand-picked and analysed manually, and thus the whole book covers only 20 cases. However, this research served as a starting point for the only (known to us) publication which tests word embedding models on Russian data to detect language change [16]. In it, gold datasets are constructed in two different ways (by applying 'pseudowords' technique and deriving changed words from [4] and [21]), and then used to evaluate some basic algorithms that had been tested before for English. [16] conclude that Kendall's $\tau$ and Jaccard distance 'remain the most relevant for detecting semantic shifts'. In the presented work, we continue experiments with word embedding models. Our results, however, are somewhat different; see **Section 6** for details. In fact, the evaluation setup in [16] seems rather convoluted; we propose more straight-forward and practically applicable approach. Additionally, we take into account the intensity of temporal semantic shifts.

---

[2] In this case the Russian National Corpus (RNC), texts from the XIX and the XX centuries.

It is important to draw the distinction between two types of semantic change: cultural shifts and linguistic drift [8]. They are linked in complex ways to two different families of data-driven methods to measure semantic change: 'global' (identifying the movement of word's vector in the whole semantic space while comparing two time periods) and 'local' (detecting variations in comparatively short lists of word's nearest semantic neighbours). After considering the decades from 1800 to 1990, [8] come to the conclusion that 'global' measures are appropriate when dealing with linguistically driven changes, whereas 'local' measures are better choice for capturing social and cultural shifts. This distinction further raises the question of how it is related to the granularity of time periods under analysis. [18] states that 'corpora with smaller time spans are useful for analysing socio-cultural semantic shifts', whereas the study of linguistically motivated semantic changes requires bigger time bins. We have used both global and local measures to compare the semantics of words as used in Russian corpora from different time periods.

## 3. Datasets

We present two gold datasets:
1. **Macro**: a list of manually chosen Russian words that have undergone semantic shifts from the pre-Soviet times through the Soviet times; this dataset is borrowed from [16].
2. **Micro**: a newly created manually annotated dataset of Russian adjectives that have undergone cultural semantic shifts in the years from 2000 to 2014.

The **Macro** dataset is a list of 215 words. 43 of them (38 nouns and 5 adjectives) are 'target' and labelled as having changed their meaning from pre-Soviet times through Soviet times, based on the linguistic research from [21] and [4]. Additionally, there are 4 fillers (words belonging to the same part of speech and the same frequency tier, randomly sampled from the Russian National Corpus) per each target word ($38 \times 4 = 152$ nouns and $5 \times 4 = 20$ adjectives). The words that have undergone semantic shifts are tagged with the class label 1, and the fillers are labelled with 0.

The **Micro** dataset (manually annotated in this work) also contains human judgements about how much the meaning of a word has shifted over a given time. However, it is limited to adjectives only[3]. To create it, we iterated over consequent pairs of yearly news texts corpora from 2000 to 2014 (2000 — 2001, ..., 2013 — 2014), and the word embedding models trained on them (see more in Section 5). To 'seed' the dataset with possible candidates for diachronic semantic shifts, for each pair of years we produced

---

[3] Adjectives are less studied in diachronic research, but at the same time present an interesting case of words less susceptible to noisy fluctuations of meaning. Note that the corresponding datasets for nouns or verbs can be easily constructed following the same workflow. We leave this for future work.

10 adjectives which shifted most according to the Global Anchors algorithm [29] (see more on it in **Section 6**). This gave us 140 adjectives with supposedly high ratio of real shifts. Then, for each of these 'seed samples', one randomly sampled adjective from the same corpus and the same frequency tier was added to the dataset as a filler. Additionally, about 20 of the seed samples were manually replaced by random fillers, mostly because of PoS tagging errors.

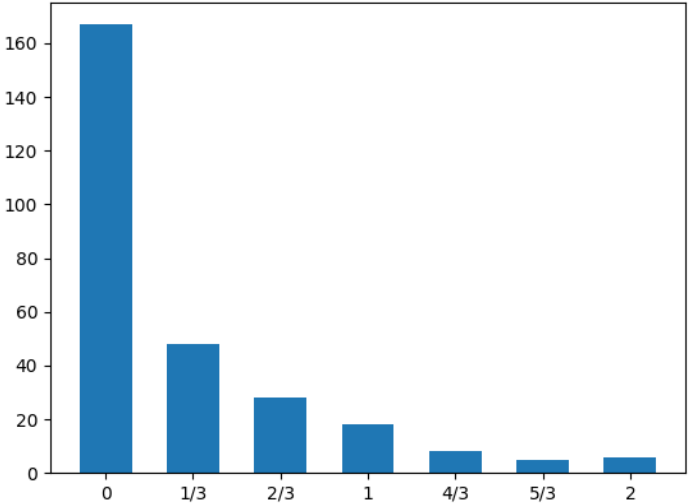

**Figure 1:** Distribution of average scores in the '**Micro**' dataset

3 human annotators independently labelled the resulting dataset of 280 adjectives from 14 year pairs. To help making the decision, each annotator was provided with a random sample of sentences containing the given adjective in the news texts from both years in the current pair. Note that during the annotation, there was no difference between the seed words and the fillers: annotators made their decision without knowing the source of adjectives. Each adjective was annotated with one of three labels:

- 0 (meaning not changed)
- 1 (meaning somewhat changed)
- 2 (meaning significantly changed)

After the first annotation round, there was a brief reconciliation, where all 3 annotators discussed the strongest disagreement cases (about 15 words out of total 280). After this reconciliation, the inter-rater agreement as measured by Krippendorff's Alpha [15] was 0.62. Considering the complexity and ambiguity of the task at hand, we believe this score to be quite high[4].

---

[4] The dataset is available online at https://github.com/wadimiusz/diachrony_for_russian

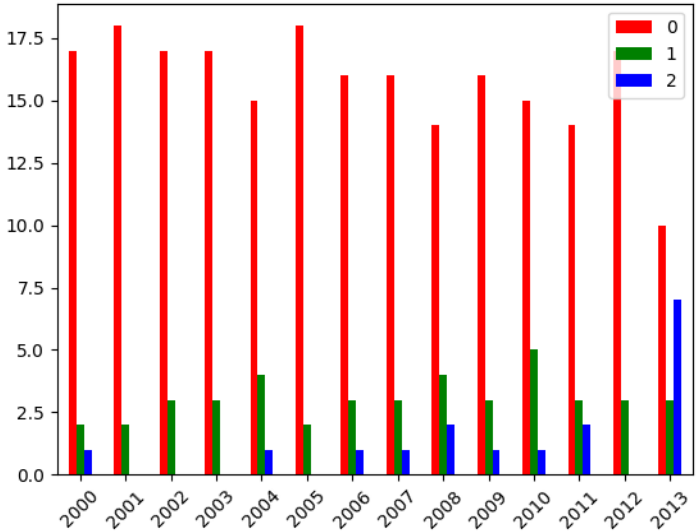

**Figure 2:** Distribution of class labels in the '**Micro**' dataset by years

The average score for each sample (over 3 annotators) could be $0, \frac{1}{3}, ..., \frac{5}{3}, 2$, seven options overall. The resulting quantized class of each word was the arithmetic mean of the three scores rounded to the nearest integer. The distribution of the mean scores over the whole dataset (before quantization into 3 classes) is shown in **Figure 1**, and the yearly distribution of class labels is shown in **Figure 2**.

**Table 1:** Socio-cultural semantic shifts in adjectives in 2014, as compared to 2013 (excerpts)

| Class | Adjective | English translation |
|---|---|---|
| 2 | крымский | 'Crimean' |
| 2 | приёмный | '1) adopted; 2) something receiving' |
| 2 | луганский | 'of Luhansk' |
| 1 | правый | '1) right; 2) right-wing' |
| 1 | кипрский | 'Cyprian, Cypriot' |
| 0 | серый | 'gray' |
| 0 | балетный | 'of ballet' |

**Table 1** contains examples of the adjectives from the 2013–2014 year pair (including two adjectives with stable meaning: 'серый' and 'балетный'). Many of the shifted adjectives are related to the Ukrainian events of 2014 that had an enormous effect on Russian news texts. For example, in 2014, the 'крымский' and 'луганский' toponyms (previously mostly associated with quiet provincial regions or seaside resorts) are almost exclusively mentioned in the context of the Russian-Ukrainian military conflict, particularly related to the annexation of the Crimea and the appearance of the self-proclaimed Luhansk People's Republic.[5]

Although geographically motivated adjectives, like the ones from the previous paragraph, are quite specific in nature (their usage can easily change due to events occurring in the corresponding locations), it still seems reasonable to include them in the dataset, since, as mentioned above, we take into account the extra-linguistic causes of semantic shifts. Moreover, the dataset contains adjectives of many different types anyway, and we believe that the toponymic ones fit well into this diversity.

The changes are not limited to this: in 2013, the word 'приёмный' was most frequently used in the sense of '*adopted*', because of very active public discussion about the law prohibiting American citizens from adopting Russian orphan children (it was enacted on January 1, 2013). However, in 2014 this topic almost disappeared from the news discourse, and the usage of 'приёмный' adjective returned to its sense of '*something receiving*' (there is also a related noun 'приёмная' '*reception office*'). Both senses had existed long before: in this case, the diachronic semantic shift consists of significant changes in the balance between these senses (a secondary sense stepping forward for social and cultural reasons, and then 'moving backwards' again).

A few adjectives were annotated as belonging to the 1 class, since the assessors believed that some changes did occur to the cultural context around these words, but they were not as significant as those described in the previous paragraph. One example is 'правый' which in 2014 acquired heavy '*right-wing*' associations with the Ukrainian nationalistic movement 'Правый сектор' (literally, '*Right sector*'), but not enough to seriously decrease its usage in the primary '*right*' sense.

Another example is 'кипрский'. In 2014 it became associated with the 'offshore' concept slightly stronger than earlier in Russian media space. This was due to the number of corruption investigations involving high-ranking Russian officials and deoffshorization initiatives that were proposed by the Russian Ministry of Finance. In 2014, 'кипрский' was often used as a generalised image, an epitome of offshore company (when no particular company is implied), as in this example:

'На заседании в Госдуме директор по маркетингу Games Operation Division (Mail. Ru Group) Михаил Кочергин назвал World of Tanks «кипрской, офшорной» игрой.' (*At a meeting in the State Duma Mikhail Kochergin, head of marketing, Games Operation Division (Mail.Ru Group), called World of Tanks a 'Cyprian, offshore' game.*)

However, it was not an entirely new sense, and that was the reason for labelling this adjective as 1. It is also worth mentioning that the inter-rater agreement was quite low for this particular adjective (the three scores were '0', '1' and '2'). The **Micro** dataset preserves the scores assigned by each assessor, so that it is possible, if desired, to use only adjectives with high inter-rater agreement.

---

[5] Of course, the dictionary definitions for these adjectives ('*related to the Crimea*' and '*related to Luhansk*' correspondingly) still remained the same. But as already noted before, distributionally, contexts actually form the meaning (including typical connotations). From this point of view, these words have certainly significantly shifted their positions in the imaginary semantic space of Russian.

Tracing cultural diachronic semantic shifts in Russian using word embeddings

## 4. Employed corpora

In accordance with the datasets described in Section 3, we employ two sets of Russian corpora:

1. '**Macro**': less granular corpora, extracted from the main body of the Russian National Corpus (RNC),[6] 3 time bins overall:
   - texts produced before 1917 (**pre-Soviet**, 75 million tokens),
   - texts produced in 1918–1990 (**Soviet**, 96 million tokens ),
   - texts produced after 1991 (**post-Soviet**, 85 million tokens).

2. '**Micro**': more granular corpora extracted from the RNC newspaper subcorpus and the *lenta.ru* dataset (14 time bins overall):
   - news texts produced in 2000,
   - news texts produced in 2001,
   - ...
   - news texts produced in 2014.

The '**Macro**' set of corpora (first used in [16]) is aimed at tracing the 'long-term' diachronic semantic shifts, while the '**Micro**' set of corpora is supposed to illustrate cases of 'short-term' cultural shifts heavily influenced by the current social and political events.

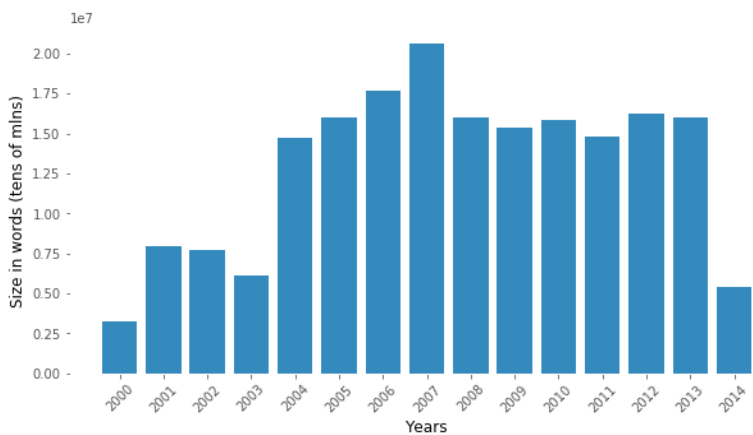

**Figure 3:** '**Micro**' corpora sizes per year

This second corpus collection (used here for the first time) features Russian news texts and comprises the whole RNC newspaper corpus[7] and the *lenta.ru* dataset texts[8] from the year 2000 up to 2014. The *lenta.ru* dataset additionally contains texts produced in the years up to 2018, but we decided to stick to the 2000–2014 time span covered by the RNC newspaper corpus, to preserve the diversity.

---

[6] http://ruscorpora.ru/en/

[7] http://ruscorpora.ru/search-paper.html

[8] https://github.com/yutkin/lenta.ru-news-dataset

The RNC newspaper corpus includes articles from the mass media of the 2000s; there are seven newspapers represented in approximately equal shares: 'Izvestia', 'Trud', 'Komsomolskaya Pravda', 'Sovetskiy Sport', 'RBC', 'RIA News', 'New Region'. The *lenta.ru* part consists of news articles crawled from one of the largest Russian news web sites. There are almost 194 million words in the whole '**Micro**' corpus, with about 12 million words per a yearly corpus on average. **Figure 3** shows yearly corpora sizes.

All the texts in the aforementioned corpora were lemmatized and tagged using a *UDPipe 1.2* model[9] trained on the Russian Universal Dependencies SynTagRus treebank [5]. Sequences of proper names immediately following each other were merged together if agreeing in case and number ('Александр_PROPN Сергеевич_PROPN' was transformed into 'Александр::Сергеевич_PROPN'). After that, lemmas were lowercased and words belonging to the functional parts of speech removed, to make it easier to train the word embedding models, described in the next section.

## 5. Word embedding models

All the approaches to trace semantic shifts employed in **Section 6** use distributional word embeddings [2, 6]. Word embeddings represent the meaning of words as dense vectors learned from lexical co-occurrences in the training corpora. In particular, we trained CBOW models [20] on all 14 + 3 = 17 tagged and lemmatized corpora described in **Section 4**, in two variations:

1. **static** models: trained separately on the corresponding time bins (years in the '**Micro**' corpora and larger time periods in the '**Macro**' corpora);

2. **incremental** models: the model $M_{n+1}$ for the year $n + 1$ is initialised with the vectors trained from the previous year model $M_n$ (essentially, this means simply continuing the training of the very first time bin model: the word vectors are updated using the co-occurrence signal from the new textual data).

Incremental models were shown to perform better for tracing temporal semantic shifts in some setups for English [17], and that was the reason we decided to test this mode for Russian data as well. Note that it also presupposes incrementally updating the vocabulary of the models.

All the models were trained with the vector dimensionality 300, context window size of 5 words to the left and 5 words to the right, for 10 epochs, with no downsampling. For the '**Macro**' models we ignored the words with corpus frequency less than 10, and for the '**Micro**' models, this parameter was set to 5, due to the corpora being significantly smaller.

---

[9] Note that we re-tagged the '**Macro**' corpora, instead of using them as they were presented in [16] (tagged with Mystem).

## 6. Experiments

In this section, we evaluate several existing approaches to semantic shifts detection on the Russian data presented above. Note that these experiments do not attempt to completely solve the problem at hand: the algorithms are very basic and intended only to provide solid baselines for further research in the area. In essence, all of the evaluated systems are implemented as logistic regression classifiers, taking as an input two word embedding models and one of the features enumerated below, and returning the semantic shift class of a word:

1. cosine distance between second order word similarity vectors in two models, using the **Global Anchors** algorithm [29];
2. cosine distance between embeddings of the word after aligning two models using the **orthogonal Procrustes transformation** [9];
3. **Jaccard distance** between top $n$ neighbour lists of one and the same word in two models [11];
4. **Kendall's** $\tau$ between top $n$ neighbour lists of one and the same word in two models [13];
5. **combined**: all the aforementioned metrics used as input features of a logistic regression.

**Procrustes alignment** is an SVD-based orthogonal transformation used to as closely as possible 'align' one embedding space to another (for example, a model trained on the 2011 corpus and the model trained on the 2012 corpus). After this, one can calculate cosine distances between word embeddings from different models, as if they were trained together. The **Global Anchors** algorithm measures how much the lists of word cosine similarities to other 'anchor words' are different between two models. In the case of *global* anchors, these lists are simply full intersection of two models' vocabularies, and thus it allows to analyse how the word position has changed related to all other words in the model.

**Jaccard distance** is a metrics used to measure the diversity between samples (here, samples are lists of a word's nearest neighbours in a given model by cosine similarity). It is defined as the difference of the samples' union and intersection sizes divided by the size of the union of two samples. **Kendall's** $\tau$ coefficient measures the rank correlation of intersections of two words' neighbours' lists. Its main difference in comparison to **Jaccard distance** when applied to our task, is that it pays attention to the relative order of $n$ nearest neighbours in two models, not only the size of their intersection.

In the terms presented in [8], we can classify Jaccard distance and Kendall's $\tau$ metrics as *local* neighbourhood measures, while the orthogonal Procrustes transformation and Global Anchors are *global* measures. The **combined** approach merges them all to find out if there are any complementary aspects in which these metrics can help each other. Because of space limitations, we do not describe the employed algorithms themselves in details. We refer the interested reader to the corresponding sources or to the survey in [18].

The 5 methods enumerated above were evaluated on both '**Micro**' and '**Macro**' datasets and the corresponding sets of diachronic word embedding models. For each method, the evaluation workflow was as follows. For each word in the current dataset, we calculated the numerical degree of its semantic shift between two models according to the current method. In the '**Macro**' dataset, these two models were always pre-Soviet and Soviet. In the '**Micro**' dataset, each word belongs to a particular sequential pair of years (e.g., 2005 and 2006), so the corresponding embedding models were used. Logistic regression classifiers were trained, using the annotated classes in the datasets as gold labels, and the calculated semantic shift degrees as features (thus, the classifiers were trained on 4 features with the **combined** method and on only 1 feature in all other cases). The '**Macro**' dataset presents a binary classification problem (shift or not shift), while the '**Micro**' dataset is a more granular ternary classification task (not shifted, somewhat shifted or significantly shifted).

These classifiers were then evaluated using stratified 9-fold cross-validation[10]. We report macro-averaged F1 score, since our class distribution is very imbalanced, but all classes are equally important. All the methods were tested both on static and incremental word embedding models, as described in Section 5.

Tables 2 and 3 present the results of the evaluation experiments with both datasets. As expected, the binary task of the '**Macro**' dataset turned out to be consistently easier for the systems under evaluation. Among the tested methods, the 'global' ones (**Global Anchors** and **Procrustes** alignment) always outperform the 'local' approaches (**Jaccard distance** and **Kendall's** $\tau$). **Kendall's** $\tau$ performed worst of all, often ending up lower than random choice. The advantage of the global methods is most clear in the '**Micro**' dataset. This emphasises the importance of taking into account the word 'trajectory' in relation to the whole vector space, when trying to detect slight social and cultural changes (while for more profound semantic shifts, it is often enough to compare the nearest neighbours' lists). Also, in the more vague ternary problem of the '**Micro**' dataset, it pays off to take into account signals from several methods: the **combined** approach yields the best performance overall. At the same time, when making strict choice between two well-defined classes in the '**Macro**' dataset, single cosine distance between word vectors after **Procrustes alignment** seems to predict semantic shifts pretty well without any other features.

**Table 2:** Macro F1 scores, 'Macro' dataset

| Model set | Global Anchors | Procrustes | Kendall | Jaccard | combined |
|---|---|---|---|---|---|
| Static | 0.675 | **0.767** | 0.504 | 0.646 | 0.722 |
| Incremental | 0.598 | 0.681 | 0.475 | 0.576 | 0.617 |
| **Random choice** | | | | | |
| $\approx 0.5$ | | | | | |

Using embedding models trained incrementally in our case did not yield strong improvements like those reported in [16] (it should be noted that our evaluation setup is significantly different). In the '**Micro**' dataset, incremental models were indeed marginally better than the static ones for all methods except **Procrustes** and **Combined**—but these two turned out to be the best, so overall the static models definitely win, being at the same time computationally simpler. It seems that employing global methods can in many cases compensate for inherent randomness in the initial states of word embedding models.

---

[10] The number of folds was motivated by the fact that there are 18 words labelled as class 2 ('significant shifts') in the '**Micro**' dataset, so 9 folds allowed us to equally distribute these samples: 2 per each fold.

Table 3: Macro F1 scores, 'Micro' dataset

| Model set | Global Anchors | Procrustes | Kendall | Jaccard | combined |
|---|---|---|---|---|---|
| Static | 0.453 | 0.468 | 0.136 | 0.301 | **0.503** |
| Incremental | 0.462 | 0.459 | 0.194 | 0.326 | 0.442 |
| Random choice | | | | | |
| ≈ 0.33 | | | | | |

In general, the achieved scores are significantly better than random, and prove that distributional word embedding models can be successfully used to trace semantic shifts of different kinds and with time periods of different granularities in Russian language material. This provides a strong baseline for further studies employing more complex approaches. At the same time, the scores show the challenging nature of the presented datasets, which we hope will cause the NLP practitioners and researchers to dig deeper into the problem of detecting temporal semantic shifts. Fully implemented code for the evaluated algorithms is published at https://github.com/wadimiusz/diachrony_for_russian/, together with the word embedding models we used.

## 7. Conclusion

We presented test sets and evaluated baseline approaches for the task of tracing diachronic (temporal) semantic shifts in Russian. We are publishing two manually annotated datasets for this task (one entirely new and another from prior work, but re-packaged and re-structured in more coherent form). The datasets are complementary in that the first one is focused on socially and culturally determined semantic shifts occurring within time spans of years, while the second one allows to test the ability of systems to trace global changes in words' meaning occurring in the scale of centuries (in this case, comparing pre-Soviet and Soviet time periods).

Further on, we used these datasets and the word embedding models trained on the corresponding corpora to rigorously evaluate 4 well-established algorithms of tracing diachronic semantic shifts (2 of which have to the best of our knowledge never been tested on Russian), as well as the system using the output of all these methods. Overall, we found that the methods using global information outperform the ones based on local data (like comparing nearest neighbours' lists). However, it should be noted that these experiments are preliminary and are intended to only provide a solid baseline for further studies in diachronic evolution of Russian lexical semantics.

For example, one obvious direction for future work is using more features describing the words (their corpus frequencies, parts of speech, semantic classes, etc). We also plan to test more advanced distributional models like those employing continuous time variables [23] or contextualised word embeddings [22]. Finally, there exists a more difficult but more fascinating task of sense-aware semantic shifts detection (tracing particular types of shifts, like narrowing, widening, or splitting the meaning into two). We believe the presented paper provides a good starting point for such kind of research in application to Russian texts.